\def\year{2019}\relax
\documentclass[letterpaper]{article} 
\usepackage{aaai19}  
\usepackage{times}  
\usepackage{helvet}  
\usepackage{courier}  
\usepackage{url}  
\usepackage{graphicx}  
\usepackage{caption}
\usepackage{amsmath}
\usepackage{amssymb}
\usepackage{subfig}

\newcommand{\citet}[1]
{\citeauthor{#1}~\shortcite{#1}}
\newcommand{\citep}{\cite}

\title{Zero-Shot Adaptive Transfer for Conversational Language Understanding}

\author{Sungjin Lee and Rahul Jha \\
  Microsoft Corporation, Redmond, WA\\
  \texttt{\{sule,rajh\}@microsoft.com}
}

\date{}

\begin{document}
\maketitle
\begin{abstract}
Conversational agents such as Alexa and Google Assistant constantly need to increase their language understanding capabilities by adding new domains. A massive amount of labeled data is required for training each new domain. While domain adaptation approaches alleviate the annotation cost, prior approaches suffer from increased training time and suboptimal concept alignments. To tackle this, we introduce a novel Zero-Shot Adaptive Transfer method for slot tagging that utilizes the slot description for transferring reusable concepts across domains, and enjoys efficient training without any explicit concept alignments. Extensive experimentation over a dataset of 10 domains relevant to our commercial personal digital assistant shows that our model outperforms previous state-of-the-art systems by a large margin, and achieves an even higher improvement in the low data regime.

\end{abstract}

\section{Introduction}
Recently, there is a surge of excitement in adding numerous new domains to conversational agents such as Alexa, Google Assistant, Cortana and Siri to support a myriad of use cases.
However, building a slot tagger, which is a key component for natural language understanding (NLU)~\cite{tur2011spoken},  for a new domain requires massive amounts of labeled data, hindering rapid development of new skills.
To address the data-intensiveness problem, domain adaptation approaches have been successfully applied. Previous approaches are roughly categorized into two groups: data-driven approaches ~\cite{DBLP:conf/coling/KimSS16,Kim2016Domainless} and model-driven approaches~\cite{kim2017domain,jha2018bag}.

In the data-driven approach, new target models are trained by combining target domain data with relevant data from a repository of arbitrary labeled datasets using domain adaptation approaches such as feature augmentation \citep{DBLP:conf/coling/KimSS16}. A disadvantage of this approach is the increase in training time as the amount of reusable data grows. The reusable data might contain hundreds of thousands of samples, making iterative refinement prohibitive.
In contrast, the model-driven approach utilizes ``expert" models for summarizing the data for reusable slots \citep{kim2017domain,jha2018bag}. The outputs of the expert models are directly used when training new domains, allowing for faster training. A drawback of this approach is that it requires explicit concept alignments which itself is not a trivial task, potentially missing lots of reusable concepts. Additionally, it's not easy to generalize these models to new, unseen slots.

\begin{figure}
  \centering
    \begin{minipage}{.45\textwidth}\centering
\subfloat[]{
  \includegraphics[width=0.9\linewidth]{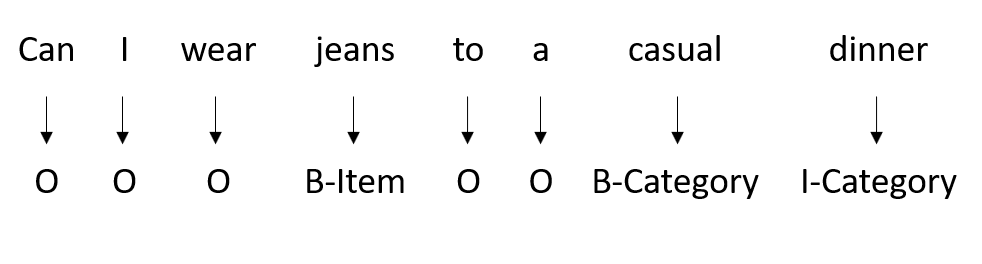}
  \label{fig:bio}}
    \end{minipage}
    \begin{minipage}{.45\textwidth}\centering    
  \subfloat[]{   
  \includegraphics[width=0.9\linewidth]{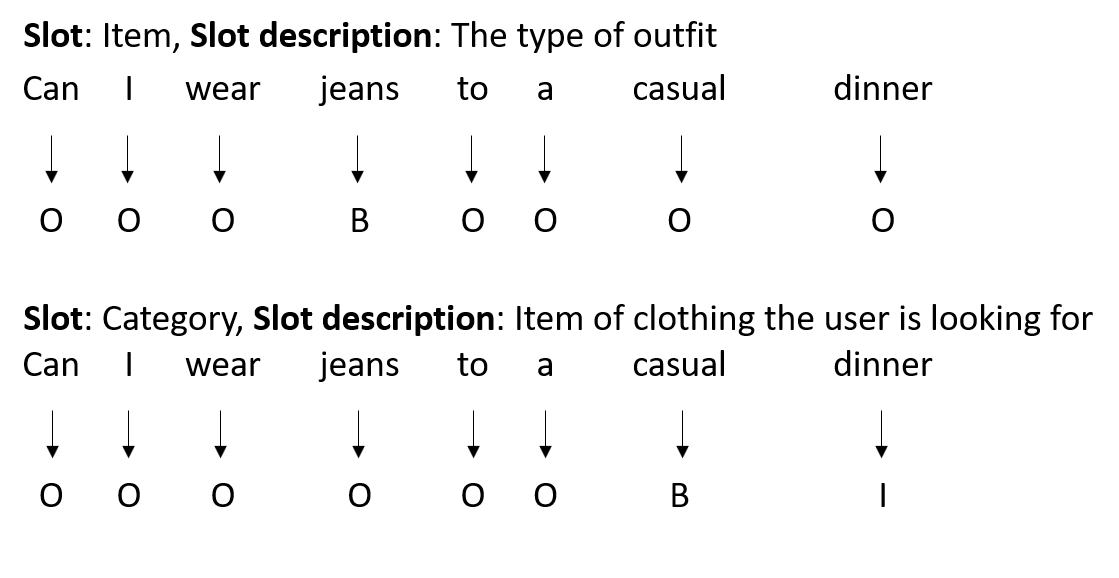}
  \label{fig:zs}}
  \end{minipage}
\caption{(a) Traditional slot tagging approaches with the BIO representation. (b) For each slot, zero-shot models independently detect spans that contain values for the slot. Detected spans are then merged to produce a final prediction.}
\label{fig:examples}
\end{figure}

\begin{figure*}[t!]
  \centering
  \includegraphics[width=0.67\linewidth]{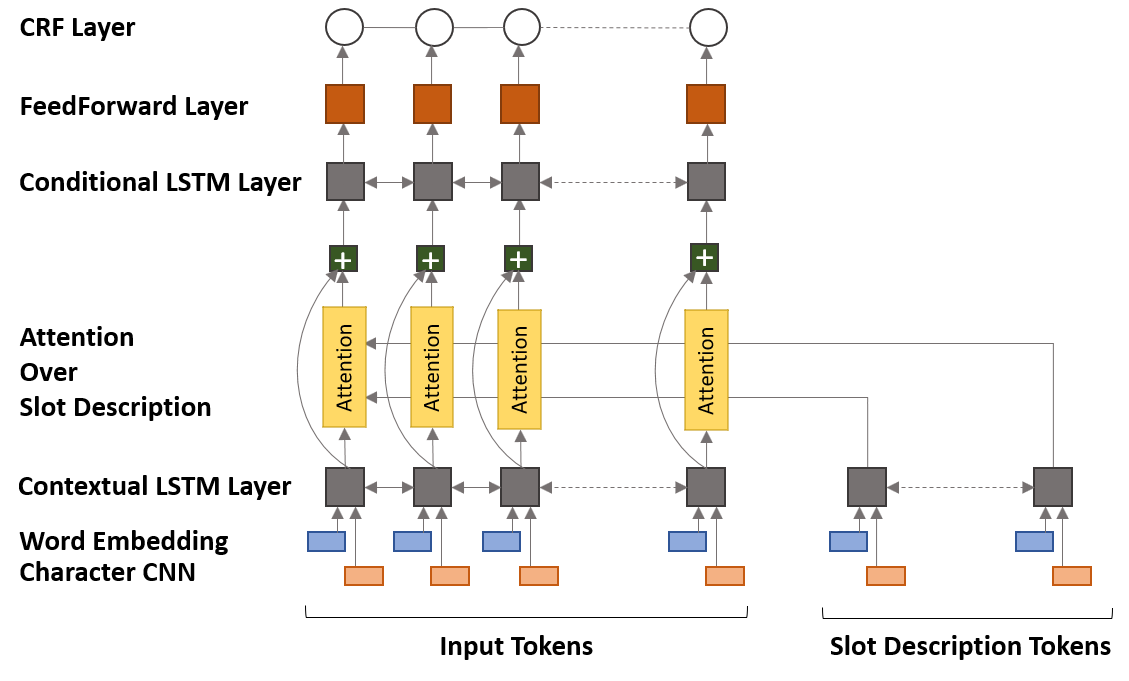}
\caption{Network architecture for the Zero-Shot Adaptive Transfer model.}
\label{fig:arch}
\end{figure*}

In this paper, we present a new domain adaptation technique for slot tagging inspired by recent advances in zero-shot learning. Traditionally, slot tagging is formulated as a sequence labeling task using the BIO representation (Figure~\ref{fig:bio}). Our approach formulates this problem as detecting spans that contain values for each slot as shown in Figure~\ref{fig:zs}. For implicit transfer of reusable concepts across domains, we represent slots in a shared latent semantic space by embedding the slot description. With the shared latent space, domain adaptation can simply be done by fine-tuning a base model, which is trained on massive data, with a handful of target domain data without any explicit concept alignments. A similar idea of utilizing zero-shot learning for slot tagging has been proven to work in semi-supervised settings~\cite{Bapna2017Toward}. Our zero-shot model architecture differs from this by adding: 1) an attention layer to produce the slot-aware representations of input words, 2) a CRF layer to better satisfy global consistency constraints, 3) character-level embeddings to incorporate morphological information. Despite its simplicity, we show that our model outperforms all existing methods including the previous zero-shot learning approach in domain adaptation settings.

We first describe our approach called {\em \textbf{Z}ero-Shot \textbf{A}daptive \textbf{T}ransfer} model (ZAT) in detail. We then describe the dataset we used for our experiments. Using this data, we conduct experiments comparing our ZAT model with a set of state-of-the-art models: Bag-of-Expert (BoE) models and their non-expert counterparts~\cite{jha2018bag}, and the Concept Tagger model~\cite{Bapna2017Toward}, showing that our model can lead to significant F1-score improvements. This is followed by an in-depth analysis of the results. We then provide a survey of related work and concluding remarks.

\section{Adaptive Transfer}
\label{meth}

Our {\em \textbf{Z}ero-Shot \textbf{A}daptive \textbf{T}ransfer} model for slot tagging is a hierarchical model with six layers (Figure~\ref{fig:arch}).  
\subsubsection{Word Embedding Layer}
Let $x_1...x_T$ and $q_1...q_J$ denote the words in the input sentence and slot description, respectively. Following~\citet{kim2014convolutional}, for every word, we obtain the character-level embedding using Convolutional Neural Networks (CNN). Characters are embedded into vectors and then get passed to the CNN. The outputs of the CNN are max-pooled to obtain a 100 dimensional vector for each word. We use pre-trained embeddings\footnote{We use 100 dimensional GloVe embeddings~\cite{pennington2014glove} for all the experiments in this study.} to obtain the word embedding of each word. We concatenate these two vectors to produce the input to the Contextual LSTM layer.

\subsubsection{Contextual LSTM Layer}
To capture the contextual meaning of words, we use a 200 dimensional bidirectional LSTM which takes the output of the previous layer as input and produces the forward and backward vectors for each word, which are then concatenated to produce a 400 dimensional vector. The outputs of this layer are two matrices: $\mathbf{X} \in \mathcal{R}^{d \times T}$ for the input sentence and $\mathbf{Q} \in \mathcal{R}^{d \times J}$ for the slot description, where {\em d} is 400. We share the same LSTM for the input sentence and the slot description.

\subsubsection{Attention Layer}
The attention layer is responsible for producing the slot-aware representations of the input words.
The inputs to the layer are contextual representations of the input sentence $\mathbf{X}$ and the slot description $\mathbf{Q}$.
For each input word $x_t$, we compute attention weights $\mathbf{a}_t \in \mathcal{R}^J$ on the slot description words:
\begin{equation*}
\mathbf{a}_{tj} = \frac{\exp(\alpha(\mathbf{x}_t,\mathbf{q}_j))}{\sum_n\exp(\alpha(\mathbf{x}_t,\mathbf{q}_n))}
\end{equation*}
where $\mathbf{q}_j$ is \textit{j}-th column vector of $\mathbf{Q}$. We choose $\alpha(\mathbf{x},\mathbf{q}) = \mathbf{w}^T[\mathbf{x};\mathbf{q};\mathbf{x}\circ\mathbf{q}]$, where $[;]$ is vector concatenation and $\circ$ is elementwise multiplication. With the attentions, we produce the slot-aware vector representations $\mathbf{G}$, where $\mathbf{G}_{:t}=\sum_j\mathbf{a}_{tj}\mathbf{q}_j$ 

\subsubsection{Conditional LSTM Layer}
We use a 200 dimensional bidirectional LSTM for the conditional layer which is responsible for capturing the interactions between the input words conditioned on the slot description.
To produce the inputs of the layer, we combine the slot-aware vector representations $\mathbf{G}$ with the contextual embeddings $\mathbf{X}$ via the elementwise summation, $\mathbf{H} = \mathbf{G} \oplus \mathbf{X}$. 

\subsubsection{Feedforward Layer}
The feedforward layer takes the output of the conditional layer, $\mathbf{H}$, as input and predicts the label scores for each word, which we denote as $\mathbf{U}$.

\subsubsection{CRF Layer}
To capture the transition behavior between labels, we use the Conditional Random Fields (CRF) layer on top of the feedforward layer. CRFs are a popular family of models that have been proven to work well in a variety of sequence tagging NLP applications~\cite{Lafferty:2001:CRF:645530.655813}.

In this study, we make predictions independently for each slot by feeding a single slot description and then obtain a final prediction by simply merging the prediction results for each slot. For example, we merge ``Find $[mexican]_{category}$ deals in seattle" and ``Find mexican eals in $[seattle]_{location}$" to produce ``Find $[mexican]_{category}$ deals in $[seattle]_{location}$." When there are conflicting spans, we select one of the spans at random. 

We initialized all LSTMs using the {\em Xavier} uniform distribution~\cite{glorot2010understanding}. We use the Adam optimizer~\cite{kingma2014adam}, with gradients computed on mini-batches of size 32 and clipped with norm value 5. The learning rate was set to $1 \times 10^{-3}$ throughout the training and all the other hyperparameters were left as suggested in \cite{kingma2014adam}. We performed early stopping based on the performance of the evaluation data to avoid overfitting.

\section{Experiments}
\subsection{Data}
\label{data}
\begin{table*}[t!]
\centering
\begin{tabular}{|l|l|l|l|}
\hline
\bf{\small Domain} & \bf{\small \#Data size} & \bf{\small \#Slots} & \bf{\small Sample Utterance} \\
\hline
{\small Fashion}          & {\small 6670} & {\small 8} & {\small Show me outfits with $[hats]_{item}$} \\
\hline
{\small Flight Status}    & {\small 10526}  & {\small 9} & {\small Status of $[Chicago~flight]_{location}$ that departed $[last~night]_{start\_time}$}\\
\hline
{\small Deals}            & {\small 28905}  & {\small 5} & {\small Find the $[best]_{rating}$ deals for $[restaurants]_{category}$}\\
\hline
{\small Purchase}         & {\small 5832} & {\small 18} & {\small Return the $[outfit]_{item}$ I purchased $[last~week]_{date}$}\\
\hline
{\small Real Estate}      & {\small 6656}  & {\small 7} & {\small Show $[houses]_{property\_type}$ $[for~rent]_{listing\_type}$ on $[Livenia~street]_{location}$}\\
\hline
{\small Shopping}         & {\small 21723}  & {\small 16} & {\small $[Gifts]_{category}$ for $[Christmas]_{keyword}$}\\
\hline
{\small Social Network}   & {\small 39323}  & {\small 21} & {\small Show $[Grace]_{username}$'s $[profile]_{media\_type}$}\\
\hline
{\small Sports}           & {\small 22437}  & {\small 21} & {\small find $[spurs]_{team\_name}$ game schedule}\\
\hline
{\small Transportation}   & {\small 202381}  & {\small 17} & {\small What's the traffic like to $[work]_{place\_type}$ }\\
\hline
{\small Travel}           & {\small 53317} & {\small 27} & {\small I need a list of $[hotels]_{accomodation\_type}$ that have $[free~kennel~services]_{amenities}$}\\
\hline
\end{tabular}
\caption{List of domains we experimented with. 80\% of the data is sampled for building the training sets, with 10\% each for dev and test sets.}
\label{tab:data}
\end{table*}

For our experiments, we collected data from a set of ten diverse domains. Table~\ref{tab:data} shows the domains along with some statistics and sample utterances. Since these are new domains for our digital assistant, we did not have enough data for these domains in our historical logs. Therefore, the data was collected using crowdsourcing from human judges. For each domain, several prompts were created to crowdsource utterances for a variety of intents. These utterances were then annotated through our standard data annotation pipeline after several iterations of measuring interannotator agreement and refining the annotation guidelines. We collected at least 5000 instances for each domain, with more data collected for some domains based on business priority.

For each of the domains, we sampled 80\% of the data as training and 10\% each as dev and test sets. Further samples of 2000, 1000, and 500 training samples were taken to compare our approach with previous methods. All samples were obtained by stratified sampling based on the annotated intents of the utterances. 


\subsection{Baseline Systems}
In order to compare our method against the state-of-the-art models, we compare against the models presented in~\cite{jha2018bag}, including the BoE models and their non-BoE variants. We also compare our method with another zero-shot model for slot tagging~\cite{Bapna2017Toward} in domain adaptation settings.

\subsubsection{LSTM}
\label{lstm}
Following~\citet{jha2018bag}, we concatenate the output of 25 dimensional character-level bidirectional LSTMs with pre-trained word embeddings to obtain morphology-sensitive embeddings. We then use a 100 dimensional word-level bidirectional LSTM layer to obtain contextualized word representations. Finally, the output of this layer is passed on to a dense feed forward layer with a softmax activation to predict the label probabilities for each word. We train using stochastic gradient descent with Adam \cite{DBLP:journals/corr/KingmaB14}. To avoid overfitting, we also apply dropout to the output of each layer, with a default dropout keep probability of 0.8. 

\subsubsection{LSTM-BoE}
The LSTM-BoE architecture is similar to the LSTM model with the exception that we use the output vectors of the word-level bidirectional LSTM layer of each expert model to obtain enriched word embeddings. Specifically, let $e_1 ... e_k \in E$ be the set of reusable expert domains. For each expert $e_j$, we train a separate LSTM model. Let $h^{e_j}_i$ be the word-level bi-directional LSTM output for expert $e_j$ on word $w_i$. When training on a target domain, for each word $w_i$, we first compute a BoE representation for this word as
$h^E = \sum_{e_i \in E} h^{e_j}_i$. The input to the word-level LSTM for word $w_i$ in the target domain is now a concatenation of the character-level LSTM outputs, the pre-trained word embedding, and the BoE representation.

Following~\citet{jha2018bag}, We use two expert domains containing reusable slots: timex and location. The timex domain consists of utterances containing the slots $date$, $time$ and $duration$. The location domain consists of utterances containing $location$, $location\_type$ and $place\_name$ slots. Both of these types of slots appear in more than 20 of a set of 40 domains developed for use in our commercial personal assistant, making them ideal candidates for reuse. Data for these domains was sampled from the input utterances from our commercial digital assistant. Each reusable domain contains about a million utterances. There is no overlap between utterances in the target domains used for our experiments and utterances in the reusable domains. The data for the reusable domains is sampled from other domains available to the digital assistant, not including our target domains. Models trained on the timex and location data have F1-scores of 96\% and 89\% respectively on test data from their respective domains.

\subsubsection{CRF}
We use a standard linear-chain CRF architecture with n-gram and context features. 
In particular, for each token, we use unigram, bigram and trigram features, along with previous and next unigrams, bigrams, and trigrams for context length of up to 3 words. We also use a skip bigram feature created by concatenating the current unigram and skip-one unigram.
We train our CRF using stochastic gradient descent with L1 regularization to prevent overfitting. The L1 coefficient was set to 0.1 and we use a learning rate of 0.1 with exponential decay for learning rate scheduling \cite{Tsuruoka:2009:SGD:1687878.1687946}.

\subsubsection{CRF-BoE}
Similar to the LSTM-BoE model, we first train a CRF model $c_j$ for each of the reusable expert domains $e_j \in E$. When training on a target domain, for every query word $w_i$, a one-hot label vector $l^j_i$ is emitted by each expert CRF model $c_j$.
The length of the label vector $l^j_i$ is the number of labels in the expert domain, with the value corresponding to the label predicted by $c_j$ for word $w_i$ set to 1, and values for all other labels set to 0. For each word, the label vectors for all the expert CRF models are concatenated and provided as features for the target domain CRF training, along with the n-gram features.

\subsubsection{CT}
For comparison with a state-of-the-art zero-shot model, we implement the concept tagger (CT)~\cite{Bapna2017Toward}. The CT model consists of a single 256 dimensional bidirectional LSTM layer that takes pre-trained word embeddings as input to produce contextual word representations. This is followed by a feed forward layer where the contextual word representations are combined with
a slot encoding to produce vectors of 128 dimensions. The slot encoding is the average vector of the word embeddings for the slot description. This feeds into another 128 dimensional bi-directional LSTM layer followed by a softmax layer that outputs the prediction for that slot. 

\subsection{Domain Adaptation using Zero-Shot Model}
For domain adaptation with zero-shot models, we first construct a joint training dataset by combining the training datasets of size 2000 from all domains except for a target domain. We then train a base model on the joint dataset. We sample input examples during training and evaluation for each slot to include both positive examples (which have the slot) and negative examples (which do not have the slot) with a ratio of 1 to 3. After the base model is trained, domain adaptation is simply done by further training the base model on varying amounts of the training data of the target domain. Note that the size of the joint dataset for each target domain is 18,000, which is dramatically smaller than millions of examples used for training expert models in the BoE approach. Furthermore, there are a lot of utterances in the joint dataset where no slots from the target domain is present.

\section{Results and Discussion}
\label{res}

\begin{table*}[h!]
\small
\center
\subfloat[]{\begin{tabular}{|l|c|c|c|c|c|c|}
  \hline
\bf{Train size} & \multicolumn{6}{|c|}{\bf{2000}} \\  
\hline
\bf{Domain} & \bf{CRF} & \bf{LSTM} & \bf{CRF-BoE} & \bf{LSTM-BoE} & \bf{CT} & \bf{ZAT}  \\
\hline
Fashion & 76.04 & 75.08 & 77.19 & 77.31 & 78.11 & \bf 81.58 \\
Flight Status & 86.46 & 89.30 & 87.91 & \bf{90.12} & 88.56 & 90.11 \\
Deals & 80.01 & 79.93 & 79.99 & 82.36 & 84.16 & \bf 84.94 \\
Purchase & 57.19 & 71.95 & 61.41 & 72.30 & 72.97 & \bf 75.33 \\
Real Estate & 91.85 & 89.47 & 91.75 & 91.01 & 91.58 & \bf 93.39 \\
Shopping & 71.96 & 73.01 & 71.45 & 72.83 & 77.06 & \bf 78.14 \\
Social Network & 81.85 & 82.15 & 81.77 & 82.24 & 79.70 & \bf 82.85 \\
Sports & 71.84 & 72.50 & 71.87 & 75.49 & 78.67 & \bf 80.83 \\
Transportation & 71.19 & 67.59 & \bf{84.94} & 79.08 & 75.54 & 80.78 \\
Travel & 62.71 & 61.50 & 67.13 & 68.20 & 72.14 & \bf 75.57 \\
\hline
Average Improvement & & $+1.14$ & $+2.43$ & $+3.98^*$ & $+4.74^*$ & $+7.24^*$ \\
\hline
\end{tabular}}

\subfloat[]{\begin{tabular}{|l|c|c|c|c|c|c|}
  \hline
\bf{Train size} & \multicolumn{6}{|c|}{\bf{1000}} \\  
\hline
\bf{Domain} & \bf{CRF} & \bf{LSTM} & \bf{CRF-BoE} & \bf{LSTM-BoE} & \bf{CT} & \bf{ZAT}  \\
\hline
Fashion & 68.73 & 74.02 & 70.98 & 71.59 & 78.90 & \bf 81.57 \\
Flight Status & 82.36 & 85.14 & 84.53 & 88.98 & 86.60 & \bf 89.88 \\
Deals & 74.60 & 70.98 & 74.13 & 74.57 & 80.69 & \bf 82.76 \\
Purchase & 52.11 & 62.05 & 53.50 & 63.91 & \bf{72.97} & 71.71 \\
Real Estate & 88.11 & 88.64 & 88.68 & 90.29 & 89.04 & \bf 91.56 \\
Shopping & 63.72 & 67.88 & 63.47 & 68.65 & 73.81 & \bf 75.52 \\
Social Network & 79.05 & 79.17 & 76.68 & 78.00 & 80.03 & \bf 84.40 \\
Sports & 63.13 & 63.75 & 63.71 & 67.25 & 73.81 & \bf 77.40 \\
Transportation & 66.45 & 60.12 & \bf{82.84} & 79.99 & 72.58 & 78.26 \\
Travel & 54.03 & 58.14 & 62.12 & 65.68 & 69.52 & \bf 69.53 \\
\hline
Average Improvement & & $+1.76$ & $+2.84$ & $+5.66^*$ & $+8.57^*$ & $+11.03^*$ \\
\hline
\end{tabular}}

\subfloat[]{\begin{tabular}{|l|c|c|c|c|c|c|}
  \hline
\bf{Train size} & \multicolumn{6}{|c|}{\bf{500}} \\  
\hline
\bf{Domain} & \bf{CRF} & \bf{LSTM} & \bf{CRF-BoE} & \bf{LSTM-BoE} & \bf{CT} & \bf{ZAT}  \\
\hline
Fashion & 62.64 & 67.55 & 66.42 & 71.59 & 73.59 & \bf 74.33 \\
Flight Status & 75.97 & 83.13 & 80.11 & 84.62 & 81.70 & \bf 86.76 \\
Deals & 64.04 & 67.55 & 67.22 & 74.24 & 77.34 & \bf 79.93 \\
Purchase & 45.19 & 57.99 & 47.76 & 60.59 & 69.33 & \bf 69.45 \\
Real Estate & 84.15 & 82.05 & 84.60 & 85.49 & 86.21 & \bf 89.14 \\
Shopping & 51.43 & 59.22 & 49.88 & 60.50 & 66.69 & \bf 69.75 \\
Social Network & 70.78 & 76.49 & 66.21 & 78.21 & 79.23 & \bf 80.39 \\
Sports & 53.29 & 55.71 & 53.85 & 63.61 & 68.20 & \bf 68.71 \\
Transportation & 60.23 & 55.18 & \bf{81.07} & 78.39 & 67.36 & 75.56 \\
Travel & 45.90 & 54.77 & 57.93 & 62.66 & 64.86 & \bf 66.34 \\
\hline
Average Improvement & & $+4.60^*$ & $+4.14$ & $+10.63^*$ & $+12.09^*$ & $+14.67^*$ \\
\hline
\end{tabular}}

\caption{F1-scores obtained by each of the six models for the 10 domains, with the highest score in each row marked as bold. Table (a), (b) and (c) report the results for 2000, 1000 and 500 training instances, respectively. The average improvement is computed over the CRF model, with the ones marked $^*$ being statistically significant with p-value $<$ 0.05.}
\label{tab:res}
\end{table*}

\subsection{Comparative Results}
Table~\ref{tab:res} shows the F1-scores~\footnote{To compute slot F1-score, we use the standard CoNLL evaluation script.} obtained by the different methods for each of the 10 domains. LSTM based models in general perform better than the CRF based models. Both the CRF-BoE and LSTM-BoE outperform the basic CRF and LSTM models. Both zero-shot models, CT and ZAT, again surpass the BoE models. 
ZAT has a statistically significant mean improvement of $4.04$, $5.37$ and $3.27$ points over LSTM-BoE with training size 500, 1000 and 2000, respectively. 
ZAT also shows a statistically significant average improvement of $2.58$, $2.44$ and $2.5$ points over CT, another zero-shot model with training size 500, 1000 and 2000, respectively.
Looking at results for individual domains, the highest improvement for BoE models are seen for \textbf{transportation} and \textbf{travel}. This can be explained by these domains having a high frequency of $timex$ and $location$ slots. But BoE models show a regression in the \textbf{shopping} domain, and a reason could be the low frequency of expert slots. In contrast, ZAT consistently outperforms non-adapted models (CRF and LSTM) by a large margin. This is because ZAT can benefit from other reusable slots than $timex$ and $location$. Though not as popular as $timex$ and $location$, slots such as $contact\_name$, $rating$, $quantity$, and $price$ appear across many domains.

\begin{figure}[h!]
  \begin{center}
    \includegraphics[width=0.45\textwidth]{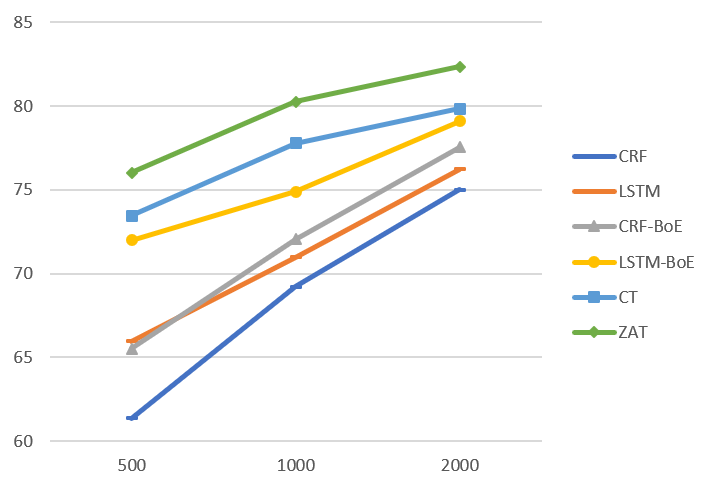}
  \end{center}
  \caption{Performance curves with varying amounts of training data for target domain.}
  \label{fig:smp_eff}
\end{figure}

We plot the averaged performances on varying amounts of training data for each target domain in Figure~\ref{fig:smp_eff}. Note that the improvements are even higher for the experiments with smaller training data. In particular, ZAT shows an improvement of $14.67$ in absolute F1-score over CRF when training with 500 instances. ZAT achieves an F1-score of 76.04\% with only 500 training instances, while even with 2000 training instances the CRF model achieves an F1-score of only 75\%. Thus the ZAT model achieves better F1-score with only one-fourth the training data.

Table~\ref{tab:zeroshot_results} shows the performances of CT and ZAT when no target domain data is available. Both models are able to achieve reasonable zero-shot performance for most domains, and ZAT shows an average improvement of $5.07$ over CT. 
\begin{table}[h!]
\small
\center
\begin{tabular}{|l|c|c|}
  \hline
\bf{Domain} & \bf{CT} & \bf{ZAT}  \\
\hline
Fashion & \bf{31.51} & 30.66 \\
Flight Status & 23.04 & \bf{25.10} \\
Deals & 37.76 & \bf{38.86} \\
Purchase & 56.34 & \bf{61.23} \\
Real Estate & 36.47 & \bf{48.63} \\
Shopping & 43.64 & \bf{50.46} \\
Social Network & 2.20 & \bf{7.22} \\
Sports & 4.49 & \bf{4.73} \\
Transportation & 39.45 & \bf{49.29} \\
Travel & 34.97 & \bf{44.34} \\
\hline
Average Improvement & & $+5.07$ \\
\hline
\end{tabular}
\caption{F1-scores with zero training instances for target domain.}
\label{tab:zeroshot_results}
\end{table}

\subsection{Model Variants}
In Table~\ref{tab:variants}, we ablate our full model by removing the CRF layer ($-CRF$) and character-level word embeddings ($-CHAR$). Without CRF, the model suffers a loss of 1\%-1.8\% points. The character-level word embeddings are also important: without this, the performance is down by 0.5\%-2.7\%. We study the impact of fine-tuning the pre-trained word embeddings ($+WEFT$). When there is no target domain data available, fine-tuning hurts performance. But, with a moderate amount of target domain data, fine-tuning improves performance.
\begin{table}[h!]
\small
\center
\begin{tabular}{|l|c|c|c|c|c|}
\hline
\bf{Model}    & \bf{0} & \bf{500} & \bf{1000} & \bf{2000}\\
\hline
ZAT          & \bf{36.05} & 76.04  & 80.26 &  82.35\\
\hline                  	     
- CRF        & 35.06 & 74.39  & 78.41 &  81.58\\
\hline                  	     
- CHAR       & 35.49 & 73.71  & 77.86 &  81.11\\
\hline                	     
+ WEFT       & 33.71 & \bf{76.52}  & \bf{80.61} &  \bf{83.09}\\
\hline
\end{tabular}
\caption{Model variants.}
\label{tab:variants}
\end{table}

\subsection{Analysis}
\begin{figure*}
  \centering
    \begin{minipage}{.24\textwidth}\centering
\subfloat[]{
  \includegraphics[width=0.85\linewidth]{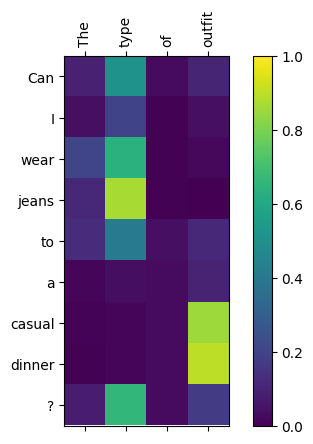}
  \label{fig:vis_sub1}}
    \end{minipage}
    \begin{minipage}{.31\textwidth}\centering    
  \subfloat[]{   
  \includegraphics[width=0.98\linewidth]{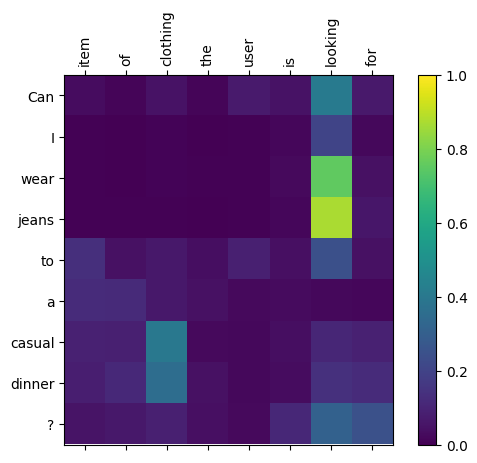}
  \label{fig:vis_sub2}}
  \end{minipage}
    \begin{minipage}{.24\textwidth}\centering    
  \subfloat[]{   
  \includegraphics[width=0.98\linewidth]{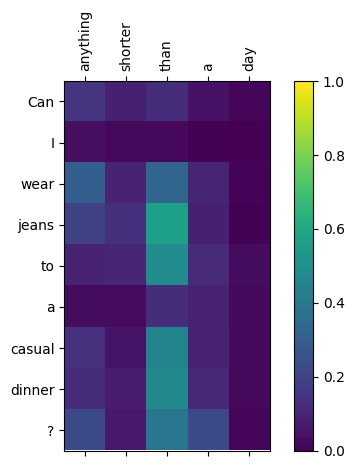}
  \label{fig:vis_sub3}}
  \end{minipage}
\caption{Visualization of attention weights for the input sentence "Can I wear jeans to a casual dinner?" with different slots: (a) category, (b) item, and (c) time.}
\label{fig:vis}
\end{figure*}

To better understand our model, in Figure~\ref{fig:vis}, we visualize the attention weights for the input sentence "Can I wear jeans to a casual dinner?" with different slots: (a) category, (b) item, and (c) time. From (a) and (b), it is clear that the attention is concentrated on the relevant words of the input and slot description. In contrast, there is no salient attention when the slot is not present in the input sentence.

\begin{figure}[h!]
  \begin{center}
    \includegraphics[width=0.4\textwidth]{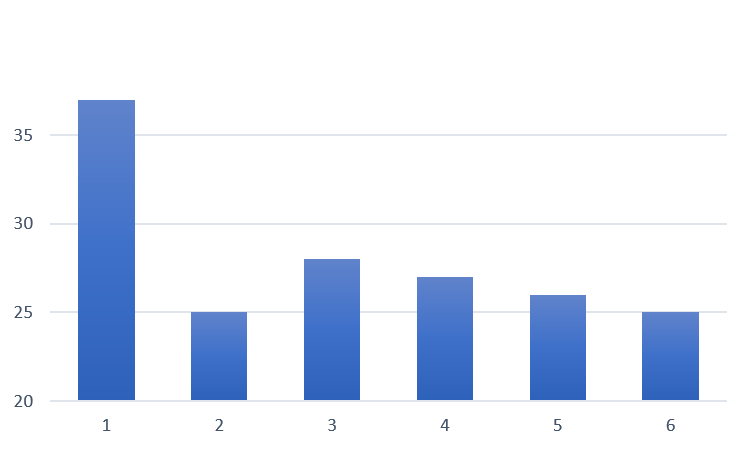}
  \end{center}
  \caption{Error rate with respect to span position}
  \label{fig:position_anal}
\end{figure}
To analyze the impact of context, we compute the error rate with respect to span start position in the input sentence. Figure~\ref{fig:position_anal} shows that error rate tends to degrade for span start positions further from the beginning. This highlights opportunities to reduce a significant amount of errors by considering previous context. 

\begin{figure}[h!]
  \begin{center}
    \includegraphics[width=0.4\textwidth]{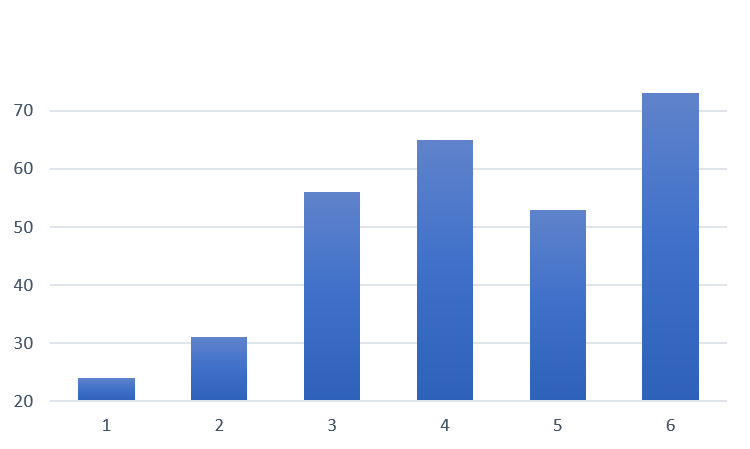}
  \end{center}
  \caption{Error rate with respect to span length}
  \label{fig:length_anal}
\end{figure}

As shown in Figure~\ref{fig:length_anal}, our model makes more errors for longer spans. This can be improved by consulting spans detected by parsers or other span-based models such as coreference resolution systems~\cite{lee2017end}.  

\begin{figure}[h!]
  \begin{center}
    \includegraphics[width=0.3\textwidth]{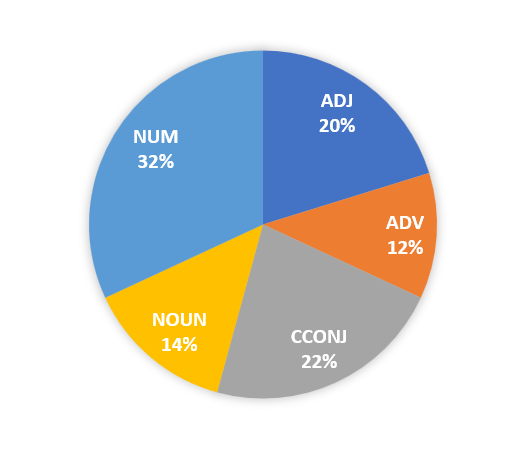}
  \end{center}
  \caption{Error rate with respect to POS tag}
  \label{fig:pos_anal}
\end{figure}

Finally, we compute the percentage of POS tags that are tied to labeling errors.~\footnote{We use spaCy for POS tagging: ~\url{https://spacy.io/}. The tag set used follows spaCy's universal POS tags.} Figure~\ref{fig:pos_anal} shows POS tags which occurs more than 10,000 times and contributes to more than 10\% of errors. It is not surprising that there are many errors for ADJ, ADV and NOUN. Our system suffers in handling conjunctive structures, for instance ``Help me find my $[black\text{ }and\text{ }tan]_{described\_as}$ $[jacket]_{item}$'', and parsing information can be helpful at enforcing structural consistencies. The NUM category is associated with a variety of concepts and diverse surface forms. Thus it is a probably good idea to have an expert model focusing on the NUM category.

\section{Related Work}
\label{rel}
A number of deep learning approaches have been applied to the problem of language understanding in recent years \citep{use-of-kernel-deep-convex-networks-and-end-to-end-learning-for-spoken-language-understanding,Mesnil2015,Celikyilmaz2018}. For a thorough overview of deep learning methods in conversational language understanding, we refer the readers to \citep{P17-5004}.

As the digital assistants increase in sophistication, an increasing number of slot models have to be trained, making scalability of these models a concern. Researchers have explored several directions for data efficient training of new models. One of the directions has been multi-task learning, where a joint model across multiple tasks and domains might be learned \citep{DBLP:journals/corr/LiuL16e,multijoint,JaechHO16}. As a recent example, \citet{Rastogi18} presented an approach for multi-task learning across the tasks of language understanding and dialog state tracking. \citet{goyal18} presented a multi-task learning approach for language understanding that consists of training a shared representation over multiple domains, with additional fine-tuning applied for new target domains by replacing the affine transform and softmax layers.   

Another direction has been domain adaptation and transfer learning methods. Early focus was on data driven adaptation techniques where data from multiple source domains was combined \citep{DBLP:conf/coling/KimSS16}. Such data-driven approaches offer model improvements at the cost of increased training time. More recently, model-driven approaches have shown success \citep{kim2017domain,jha2018bag}. These approaches follow the strategy of first training expert models on the source data, and then using the output of these models when training new target models. A benefit of these approaches over data-driven adaptation techniques is the improved training time that scales well as the number of source domains increase. 

However, both these transfer learning approaches require concept alignment to map the new labels to existing ones, and cannot generalize to unseen labels. This has led researchers to investigate zero-shot learning techniques, where a model is learned against label representations as opposed to a fixed set of labels. 

Several researchers have explored zero-shot models for domain and intent classification. \citet{Dauphin2014Zero} described a zero-shot model for domain classification of input utterances by using query click logs to learn domain label representations. \citet{Kumar2017Zero} also learn a zero-shot model for domain classification. \citet{chenzeroshot} learn a zero-shot model for intent classification using a DSSM style model for learning semantic representations for intents.

Slot tagging using zero-shot models has also been explored. \citet{Ferreira2015ZeroshotSP} presented a zero-shot approach for slot tagging based on a knowledge base and word representations learned from unlabeled data. \citet{Bapna2017Toward} also applied zero-shot learning to slot-filling by implicitly linking slot representations across domains by using the label descriptions of the slots. Our method is similar to their approach, but we use an additional attention layer to produce the slot-aware representations of input words, leading to better performance as demonstrated by our empirical results. 

More recently, zero-shot learning has also been applied to other tasks. For example, \citet{upadhyay18} apply zero-shot learning for training language understanding models for multiple languages and show good results. \citet{hadyzeroshot} presented a zero-shot model for question generation from knowledge graphs, and \citet{huangzeroshot} presented a model for zero-shot transfer learning for event extraction.

\section{Conclusion}
\label{con}
In this paper, we introduce a novel Zero-Shot Adaptive Transfer method for slot tagging that utilizes the slot description for transferring reusable concepts across domains to avoid some drawbacks of prior approaches such as increased training time and suboptimal concept alignments. 
Experiment results show that our model performs significantly better than state-of-the-art systems by a large margin of 7.24\% in absolute F1-score when training with 2000 instances per domain, and achieves an even higher improvement of 14.57\% when only 500 training instances are used. 
We provide extensive analysis of the results to shed light on future work. We plan to extend our model to consider more context and utilize exogenous resources like parsing information.



\bibliography{refs}
\bibliographystyle{aaai}

\end{document}